\documentclass{article}

\PassOptionsToPackage{numbers, compress}{natbib}
\usepackage[final]{neurips_2023}

\usepackage[utf8]{inputenc} % allow utf-8 input
\usepackage[T1]{fontenc}    % use 8-bit T1 fonts
\usepackage[pagebackref=true]{hyperref}       % hyperlinks
\usepackage{url}            % simple URL typesetting
\usepackage{booktabs}       % professional-quality tables
\usepackage{amsfonts}       % blackboard math symbols
\usepackage{nicefrac}       % compact symbols for 1/2, etc.
\usepackage{microtype}      % microtypography
\usepackage{xcolor}         % colors

\usepackage{caption}
\usepackage{subcaption}
\usepackage{xkcdcolors}
\usepackage{siunitx}
\usepackage{tikz-network}
\usepackage{fontawesome}
\usepackage{pgffor}
\usepackage{xfp}
\usepackage{wrapfig}
\usepackage{stmaryrd}
\usepackage{bm}
\usepackage{tablefootnote}

\usetikzlibrary{decorations,calligraphy}
\DeclareCaptionLabelFormat{short}{Fig.~#2}
\DeclareCaptionLabelFormat{long}{#1~#2}
\colorlet{onno}{xkcdPeriwinkleBlue}

\title{Middle-Mile Logistics Through the Lens of Goal-Conditioned Reinforcement Learning}

\author{%
  Onno Eberhard\textnormal{\textsuperscript{1,2}}\thanks{Work performed during an internship at Google Research. Correspondence: \texttt{oeberhard@tue.mpg.de}.}\\
  \And
  Thibaut Cuvelier\textnormal{\textsuperscript{3}}\\
  \And
  Michal Valko\textnormal{\textsuperscript{4}}\\
  \And
  Bruno De Backer\textnormal{\textsuperscript{3}}\\
  \AFF
  \textsuperscript{1}Max Planck Institute for Intelligent Systems, Tübingen, Germany\\
  \Aff
  \textsuperscript{2}University of Tübingen\\
  \Aff
  \textsuperscript{3}Google Research\\
  \Aff
  \textsuperscript{4}Google DeepMind\\
}

\begin{document}

\maketitle

\begin{abstract}
Middle-mile logistics describes the problem of routing parcels through a network of hubs, which are linked by a fixed set of trucks.
The main challenge comes from the finite capacity of the trucks.
The decision to allocate a parcel to a specific truck might block another parcel from using the same truck.
It is thus necessary to solve for all parcel routes simultaneously.
Exact solution methods scale poorly with the problem size and real-world instances are intractable.
Instead, we turn to reinforcement learning (RL) by rephrasing the middle-mile problem as a multi-object goal-conditioned Markov decision process.
The key ingredients of our proposed method for parcel routing are the extraction of small feature graphs from the environment state and the combination of graph neural networks with model-free RL.
There remain several open challenges and we provide an open-source implementation of the environment to encourage stronger cooperation between the reinforcement learning and logistics communities.
\end{abstract}

\hfill
\begin{minipage}[b]{0.48\textwidth}
  \centering
  \captionsetup{type=figure}
  \begin{tikzpicture}[scale=0.85]
    \Vertex[IdAsLabel, x=0, y=1]{A}
    \Vertex[IdAsLabel, x=1.5, y=2]{B}
    \Vertex[IdAsLabel, x=2, y=0]{C}
    \Vertex[IdAsLabel, x=3.5, y=0.5]{D}
    \Vertex[IdAsLabel, x=4, y=2.5]{E}
    \node[scale=1.2] (fac) at (-1.5, 2) {\faIndustry};
    \node[scale=1.5] (home) at (5, 0) {\faHome};
    \node[scale=0.8, text=xkcdDarkPink] (parcel1) at (0.55, 1) {\faGift};
    \node[scale=0.8, text=xkcdGrassGreen] (parcel2) at (2, 1.8) {\faGift};
    \Edge(A)(B)
    \Edge(A)(C)
    \Edge(B)(C)
    \Edge(B)(E)
    \Edge[label=$T_1$](C)(D)
    \Edge[label=$T_2$](D)(E)
    \draw[arrows={-Stealth[inset=0pt, scale=.7]}, color=xkcdDarkPink, shorten >=1.5pt, shorten <=-0.5pt] (parcel1) -- (D);
    \draw[arrows={-Stealth[inset=0pt, scale=.7]}, color=xkcdGrassGreen, shorten >=1.5pt, shorten <=-2pt] (parcel2) -- (D);
    \draw[arrows={-Stealth[inset=0pt, scale=.7]}, densely dashed, thick, shorten >=0.5pt, shorten <=-1pt] plot [smooth, tension=1] coordinates {(fac.south) (-1.2, 0.2) (-0.3, 0) (-0.8, 1) (-0.35, 1.1)};
    \draw[arrows={-Stealth[inset=0pt, scale=.7]}, densely dashed, thick, shorten >=-2pt] plot [smooth, tension=1] coordinates {(D.east) (4.2, 0.5) (4.6, 1.5) (home.north)};
    \draw [very thick,decorate,decoration={mirror, calligraphic brace,amplitude=5pt}] (-1.7,-0.4) -- (-0.2,-0.4) node [scale=0.7,midway,yshift=-0.5cm] {First-mile};
    \draw [very thick,decorate,decoration={mirror, calligraphic brace,amplitude=5pt}] (-0.2,-0.4) -- (3.7,-0.4) node [scale=0.7,midway,yshift=-0.5cm] {Middle-mile};
    \draw [very thick,decorate,decoration={mirror, calligraphic brace,amplitude=5pt}] (3.7,-0.4) -- (5.2,-0.4) node [scale=0.7,midway,yshift=-0.5cm] {Last-mile};
    % \draw (current bounding box.north east) -- (current bounding box.north west) -- (current bounding box.south west) -- (current bounding box.south east) -- cycle;
    % \draw[step=1cm,gray,very thin] (-2,-1) grid (6,6);
  \end{tikzpicture}
  
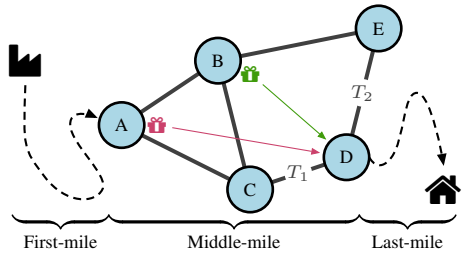
\captionof{figure}{The middle-mile logistics problem}
  \label{fig:middle-mile}
\end{minipage}
\hfill
\hfill
\begin{minipage}[b]{0.48\textwidth}
  \centering
  \captionsetup{type=figure}
  \begin{tikzpicture}[scale=0.48]
    \foreach \t in {1,...,5}{
      \node (t\t) at (-1.5, -\t) {\small $t = \t$};
      \foreach \i in {0,...,4}{
        \Vertex[size=.2, x=\i, y=-\t]{\i\t}
        \Vertex[size=.2, x=\fpeval{\i + 6}, y=-\t]{\i\t'}
      }
    }
    \node at (0, 0) {\small A};
    \node at (1, 0) {\small B};
    \node at (2, 0) {\small C};
    \node at (3, 0) {\small D};
    \node at (4, 0) {\small E};
    \node at (6, 0) {\small A};
    \node at (7, 0) {\small B};
    \node at (8, 0) {\small C};
    \node at (9, 0) {\small D};
    \node at (10, 0) {\small E};
    \Vertex[size=.2, x=0, y=-1, color=xkcdDarkPink]{P1}
    \Vertex[size=.2, x=1, y=-1, color=xkcdGrassGreen]{P2}
    \Vertex[size=.2, x=3, y=-5, label=$\star$]{G}
    \foreach \t in {1,...,4}{
      \foreach \i in {0,...,4}{
        \Edge[Direct, lw=0.2, opacity=0.3]](\i\t)(\i\fpeval{\t + 1})
      }
    }
    \Edge[Direct, lw=0.8, color=xkcdOrangish](01)(22)
    \node[color=xkcdOrangish] at (1.45, -1.35) {\small $a$};
    \Edge[Direct, lw=0.8](41)(32)
    \Edge[Direct, lw=0.8](11)(03)
    \Edge[Direct, lw=0.8](12)(23)
    \Edge[Direct, lw=0.8](12)(43)
    \Edge[Direct, lw=0.8](13)(05)
    \Edge[Direct, lw=0.8](23)(34)
    \Edge[Direct, lw=0.8](44)(35)  
    \node[align=center] at (5, -2.5) {\tiny MDP};
    \node[align=center] at (5, -3) {$\mapsto$};  
    \Vertex[size=.2, x=8, y=-2, color=xkcdDarkPink]{P1'}
    \Vertex[size=.2, x=7, y=-1, color=xkcdGrassGreen]{P2'}
    \Vertex[size=.2, x=9, y=-5, label=$\star$]{G'}
    \foreach \t in {1,...,4}{
      \foreach \i in {0,...,4}{
        \Edge[Direct, lw=0.2, opacity=0.3]](\i\t')(\i\fpeval{\t + 1}')
      }
    }
    \Edge[Direct, lw=0.8](01')(22')
    \Edge[Direct, lw=0.8](41')(32')
    \Edge[Direct, lw=0.8](11')(03')
    \Edge[Direct, lw=0.8](12')(23')
    \Edge[Direct, lw=0.8](12')(43')
    \Edge[Direct, lw=0.8](13')(05')
    \Edge[Direct, lw=0.8](23')(34')
    \Edge[Direct, lw=0.8](44')(35')  
    \Vertex[size=.2, x=-2, y=-6, color=xkcdDarkPink]{LP1}
    \node[anchor=base] at (-0.1, -6.23) {\small Red parcel};
    \Vertex[size=.2, x=2.9, y=-6, color=xkcdGrassGreen]{LP2}
    \node[anchor=base] at (5.1, -6.23) {\small Green parcel};
    \Vertex[size=.2, x=8.2, y=-6, label=$\star$]{LG}
    \node[anchor=base] (LGT) at (9.4, -6.23) {Goal};  
    % \draw (t5.south west) -- (LGT.north east) -- (LGT.south east) -- (current bounding box.south west) -- cycle;
    % \draw[step=1cm,gray,very thin] (-2,-1) grid (6,6);
  \end{tikzpicture}
  
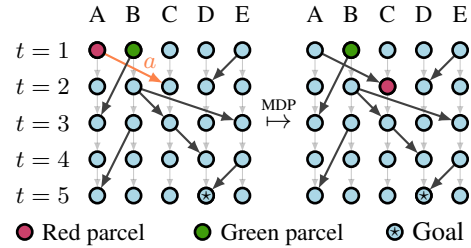
\captionof{figure}{Time-expanded graphs \& MDP dynamics}
  \label{fig:time-expanded}
\end{minipage}
\hfill\null

\section{Introduction}
The logistics of transporting a parcel is usually organized in three steps: first-mile (getting the parcel from the sender to the start hub in a transportation network), middle-mile (routing the parcel between the start and goal hubs), and last-mile (delivering the parcel from the goal hub to the receiver); see Fig.~\ref{fig:middle-mile}.
In this work, we only consider the middle-mile problem, and assume that all parcels and trucks are fixed.
The problem consists in finding a route (a collection of trucks) for each parcel such that all parcels reach their destinations on time.
It is straightforward to find such a route for a single parcel by employing a shortest path algorithm.
The real challenge lies in consolidating the parcel routes, because trucks only have finite capacity.
This is illustrated in a small example in Fig.~\ref{fig:middle-mile}.
Here, two parcels (red and green) share a common goal.
Routing the parcels individually might result in both parcels traveling via truck $T_1$.
However, if $T_1$ is too small to carry both, the delivery would fail.
A middle-mile solver has to take this into account and would route the green parcel via $T_2$ instead.

Another difficulty of the middle-mile problem is its dimension: real-world instances may include 100s of hubs, 10,000s of time steps, and 1,000,000s of parcels.
Current solution approaches mostly rely on combinations of mixed-integer programming techniques with hand-crafted heuristics~\cite{bakir-2021-snd,gendron-2016-ilph}.
While these algorithms can achieve good solutions for small instances, they fail to scale to large ones, even though these are the most common in practice.
In this work, we investigate whether scalable routing strategies can instead be learned using reinforcement learning (RL).
Reinforcement learning has been used for combinatorial optimization~\cite{bengio-2021-mlco,mazyavkina-2021-rlco}, and more specifically in similar domains such as ride hailing~\cite{qin-2021-ridehailing} and vehicle routing~\cite{nazari-2018-vrp}, which suggests that an RL approach might also be fruitful in this domain.
To our knowledge, this is the first application of RL to the middle-mile logistics problem.
\looseness=-1

\section{Method}
To represent a middle-mile problem, we use the \textit{time-expanded} logistics network, which includes all information about the individual trucks and parcels.
This representation makes the temporal aspect of the routing problem explicit, and is shown on the left of Fig.~\ref{fig:time-expanded} for the small example from above.
This graph can be used to turn the logistics problem into a Markov decision process (MDP), with the graph itself being the state.
In this MDP, an action is the assignment of a parcel to one of the available trucks.
The MDP dynamics are also illustrated in Fig.~\ref{fig:time-expanded}, which shows a state-action pair being mapped to a new state.
An alternative MDP definition might take just the parcel locations as the state.
However, this state representation would not allow for generalization to new logistics networks, so we use the complete graph as the state.
We consider staying at the same hub for one time step as using a ``virtual truck'' with infinite capacity (light gray edges in Fig.~\ref{fig:time-expanded}).
A transition changes the state in two ways: the parcel location shifts to where the truck takes it (in time and space), and the truck's capacity shrinks by the weight of the parcel.
More details on the MDP are provided in Sec.~\ref{sec:mdp}.

We define a simple reward function: $1$ if a transition results in a successful delivery and $0$ otherwise.
The value of a state under this reward function is thus the expected number of parcels that will be delivered in the future when starting at this state and using the given policy.
This complete MDP definition allows us to use any reinforcement learning method.
A value-based method (such as fitted value iteration) might seem most appropriate, as we have access to the environment dynamics.
However, due to the difficulty of learning an accurate global value function, we decided on a policy search approach instead.
In particular, we use a custom version of the proximal policy optimization (PPO) algorithm~\cite{schulman-2017-proximal}, implemented in JAX~\cite{bradbury-2018-jax}, designed to work in this setting, where the number of possible actions depends on the state.

As our states are represented as graphs, we chose to use graph neural network (GNN) function approximation, based on the GraphNet framework~\cite{battaglia-2018-relational}.
The main difficulty with this approach is the size of the graphs.
To propagate information through the network with message-passing GNNs, the running time grows at least linearly with the graph size.
Real-world logistics problems can be very large, but even with the smaller graphs in our experiments, GNN processing becomes prohibitive.
We employ two different strategies to deal with this issue.
First, we prune the state by removing nodes and edges that are irrelevant to the problem (see Sec.~\ref{sec:pruning}).
Second, we remove all nodes and edges outside a small radius from the parcel position and goal, only leaving a small ``feature graph.''
Because this potentially throws away important information, we include additional node and edge features, such as distance-to-goal heuristics (see Sec.~\ref{sec:feature-graphs}).

Both actor and critic (which don't share weights in our implementation) take the form of an encode-process-decode architecture, implemented using the Jraph library~\cite{godwin-2020-jraph}, and take the feature graph as input.
The encoder is a GraphNet block that maps the node and edge features into a latent space using multi-layer perceptron (MLP) node and edge update functions.
The processor is a GraphNet block that updates the latent node and edge features, also using MLP node and edge update functions.
The processor update has a residual connection, such that the new node and edge features are \(\bm x' = \bm y + \bm x\), where \(\bm x\) are the previous features and \(\bm y\) is the output of the MLP.
The processor is applied several times (with the same weights) before the decoder is applied.
The decoder is a GraphNet block that only updates the edge features, which it maps from the latent dimension to the real numbers using an MLP.
Finally, all but the edges corresponding to valid actions are removed using masking techniques.
In the critic, the remaining values then represent the Q-value estimates \(q_\phi(s, a)\).
In the actor, they represent policy logits, to which a softmax function with an inverse temperature \(\alpha\) is applied to yield a policy $\pi_{\theta}(a \mid s)$ with Boltzmann exploration.
The advantage of a state-action pair \((s, a)\) is then estimated as \(q_\phi(s, a) - \sum_{a'} \pi_\theta(a' \mid s) q_\phi(s, a')\).

\section{Experiments}
\captionsetup[figure]{labelfont=bf, labelformat=long}
\begin{wrapfigure}{r}{0.5\textwidth}
  \vspace{-1.4em}
  \begin{center}
    \includegraphics{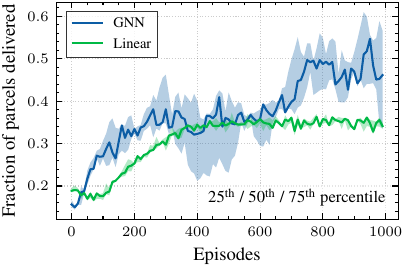}
  \end{center}
  \vspace{-0.5em}
  \caption{PPO learning curves. Shown are the returns from exploratory rollouts (5 seeds each).}
  \label{fig:ppo}
  \vspace{-1.2em}
\end{wrapfigure}
We test our method on a small logistics network with 10 hubs, 50 time steps, and 200 parcels.
The learning curves of PPO in this environment are shown in Fig.~\ref{fig:ppo}.
For each episode, we sample a new logistics network with new parcels.
Thus, it can be seen that the method learns the general task of routing parcels through logistics networks, not just a routing strategy for a specific network.

In Fig.~\ref{fig:ppo}, we compare the GNN parameterization of the actor and critic explained above with a simpler linear method.
Given a vector representation of a state-action tuple \(\bm x : \mathcal S \times \mathcal A \to \mathbb R^d\), the linearly parameterized policy and Q-function take the forms
\begin{align*}
    \pi_{\bm\theta}(a \mid s) \propto \exp\left\{\alpha \bm\theta^\top \bm x(s, a)\right\} \quad \text{and} \quad q_{\bm\phi}(s, a) = \bm\phi^\top \bm x(s, a),
\end{align*}
where \(\bm\phi, \bm\theta \in \mathbb R^d\) are the learned parameters and \(\alpha\) is an inverse temperature coefficient controlling the amount of Boltzmann exploration.
The feature vector \(\bm x(s, a)\) is constructed from the 1-step feature graph (see Sec.~\ref{sec:feature-graphs}) of the state \(s\) by concatenating the edge features of truck \(a\) and the node features of the receiving node of that truck.
It can be seen that the GNN approach performs better than the linear method.

\begin{figure}[tb]
    \centering
    \includegraphics{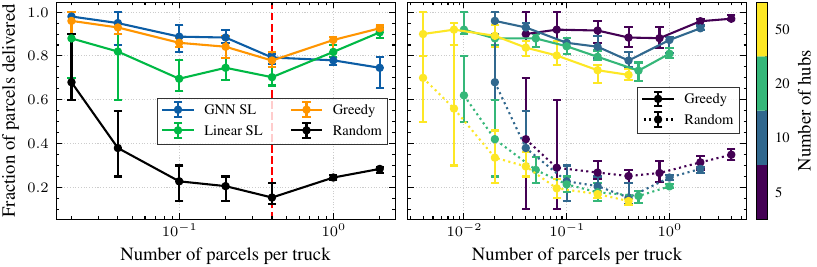}
    \caption{Performances (min / max / mean of 5 random seeds) of different policies on a 50-time step (left: 10-hub) network, as the number of parcels is changed. Left: Comparison of policies trained with supervised learning, a simple greedy policy, and a uniformly random policy. The supervised learning policies were only trained on data from 200-parcel networks (red dashed line). Right: The difficulty of the middle-mile problem depends on many factors: the number of hubs, the number of timesteps, and the number of parcels. If there are too few or too many parcels in the network, the problem becomes easier.}
    \label{fig:hubs-supervised}
\end{figure}

% \subsection{Supervised learning}\label{sec:supervised}
In addition to the reinforcement learning experiments, we also trained linear and GNN policies using supervised learning.
As we explain in Sec.~\ref{sec:initialization}, when initializing the middle-mile MDP state, the parcels are placed by sampling complete parcel routes.
When taken together, these routes form a valid solution to the middle-mile problem.
Thus, we can use the actions from these parcel routes as targets in a supervised learning setup.
The left plot in Fig.~\ref{fig:hubs-supervised} shows the performances of linear and GNN policies when trained in this way.
The policies were trained on data collected from 200-parcel networks, indicated by the red line in the plot.
It can be seen that both the linear and the GNN policies generalize well to different numbers of parcels at inference time.
We also show the performance of a greedy policy that always selects the truck whose destination hub has the lowest resistance distance to the goal hub (see Sec.~\ref{sec:feature-graphs}), as well as the performance of a uniformly random policy.

The difficulty of the middle-mile problem depends greatly on how packed the network is with parcels.
If there are only very few parcels, then the problem becomes easier because the middle-mile problem can be decomposed into individual shortest-path problems.
If there are so many parcels that they barely fit in the available number of trucks, then sampling interesting parcel routes, i.e.\ ones that use few virtual trucks, becomes more difficult.
As a result, solving these problems becomes easier, as the routes are more straightforward, sometimes not even changing hub location once.
This can be seen in the left plot of Fig.~\ref{fig:hubs-supervised} but becomes even clearer in the right plot, where we analyze how the performance of the greedy and random policies depends on the number of hubs, which in this case also equals the number of trucks per time step (see Sec.~\ref{sec:mdp}).

\section{Conclusion}
Comparing Figs.~\ref{fig:ppo} and \ref{fig:hubs-supervised}, it can be seen that supervised learning, and even the simple greedy baseline, both outperform the more complex reinforcement learning approach.
If this is the case, why should we use reinforcement learning at all?
The case against the greedy policy is simple: it isn't enough.
Just like the supervised learning and RL approaches, the greedy policy does not solve the middle-mile problem.
However, unlike the learning-based methods, there is no way to improve the greedy method other than by constructing more elaborate heuristics.
This is exactly the current situation of middle-mile logistics and the reason why we turned to learning-based methods in the first place.
The reason why supervised learning is not a good option, at least not as good as reinforcement learning, is more subtle.
It has to do with the data that is used to train the policies.
This data does \textit{not} consist of optimal trajectories.
It consists of the sampled parcel routes, which are not designed to be optimal.
With the 0-1 reward we currently consider this distinction is not very clear, but when more realistic cost functions are used that take the real costs of shipping or dropping parcels into account, there is no easy way to construct a dataset for supervised learning containing optimal labels for this objective.

Our results show that reinforcement learning methods can be used in this context, though there is still a long way to go for good performance.
The presented approach generalizes to stochastic dynamics (e.g. parcels arriving gradually in the network), to more realistic cost functions (taking the actual costs of shipping or dropping parcels into account), as well as to vector-valued parcel weights (e.g. weight and size).
The model-free method we present ignores the fact that we have access to the dynamics; a better approach might include model-based search methods such as Monte Carlo tree search.
As mentioned above, the value of a state is the expected number of future deliveries.
This is very difficult to estimate from the small feature graphs, as it is impossible to know how many other parcels there are in the complete state.
Thus, another improvement may come from \textit{direct advantage estimation}~\cite{pan-2022-direct} to replace the Q-critic.
It seems reasonable that the advantage, as the expected number of additional parcels that would be delivered by choosing a certain action over following the policy, can be accurately estimated even from small feature graphs.

Our main contribution is our open-source implementation of the middle-mile logistics environment (see Sec.~\ref{sec:mdp}), which we provide online (\url{https://github.com/google-research/laurel}) in the hope that it will open the door to an important real-world application of goal-conditioned RL.

\section*{Acknowledgments}
We want to thank Jessica Hamrick, Sephora Madjiheurem, Rémi Munos, Luis Piloto, and Ondrej Sykora for helpful discussions about the project, and the International Max Planck Research School for Intelligent Systems (IMPRS-IS) for supporting Onno Eberhard.

\bibliographystyle{abbrvnat}
\bibliography{bib}

\begin{thebibliography}{16}
\providecommand{\natexlab}[1]{#1}
\providecommand{\url}[1]{\texttt{#1}}
\expandafter\ifx\csname urlstyle\endcsname\relax
  \providecommand{\doi}[1]{doi: #1}\else
  \providecommand{\doi}{doi: \begingroup \urlstyle{rm}\Url}\fi

\bibitem[Albert and Barab{\'{a}}si(2000)]{albert-2000-topology}
R.~Albert and A.-L. Barab{\'{a}}si.
\newblock Topology of evolving networks: Local events and universality.
\newblock \emph{Physical Review Letters}, 85\penalty0 (24):\penalty0
  5234--5237, Dec. 2000.
\newblock URL \url{https://doi.org/10.1103/physrevlett.85.5234}.

\bibitem[Bakir et~al.(2021)Bakir, Erera, and Savelsbergh]{bakir-2021-snd}
I.~Bakir, A.~Erera, and M.~Savelsbergh.
\newblock Motor carrier service network design.
\newblock In \emph{Network Design with Applications to Transportation and
  Logistics}, pages 427--467. {Springer International Publishing}, 2021.
\newblock URL \url{https://doi.org/10.1007/978-3-030-64018-7_14}.

\bibitem[Battaglia et~al.(2018)Battaglia, Hamrick, Bapst, Sanchez{-}Gonzalez,
  Zambaldi, Malinowski, Tacchetti, Raposo, Santoro, Faulkner,
  G{\"{u}}l{\c{c}}ehre, Song, Ballard, Gilmer, Dahl, Vaswani, Allen, Nash,
  Langston, Dyer, Heess, Wierstra, Kohli, Botvinick, Vinyals, Li, and
  Pascanu]{battaglia-2018-relational}
P.~W. Battaglia, J.~B. Hamrick, V.~Bapst, A.~Sanchez{-}Gonzalez, V.~F.
  Zambaldi, M.~Malinowski, A.~Tacchetti, D.~Raposo, A.~Santoro, R.~Faulkner,
  {\c{C}}.~G{\"{u}}l{\c{c}}ehre, H.~F. Song, A.~J. Ballard, J.~Gilmer, G.~E.
  Dahl, A.~Vaswani, K.~R. Allen, C.~Nash, V.~Langston, C.~Dyer, N.~Heess,
  D.~Wierstra, P.~Kohli, M.~M. Botvinick, O.~Vinyals, Y.~Li, and R.~Pascanu.
\newblock Relational inductive biases, deep learning, and graph networks.
\newblock \emph{CoRR}, abs/1806.01261, 2018.
\newblock URL \url{https://arxiv.org/abs/1806.01261}.

\bibitem[Bengio et~al.(2021)Bengio, Lodi, and Prouvost]{bengio-2021-mlco}
Y.~Bengio, A.~Lodi, and A.~Prouvost.
\newblock Machine learning for combinatorial optimization: {A} methodological
  tour d'horizon.
\newblock \emph{European Journal of Operational Research}, 290\penalty0
  (2):\penalty0 405--421, 2021.
\newblock URL \url{https://doi.org/10.1016/j.ejor.2020.07.063}.

\bibitem[Bradbury et~al.(2018)Bradbury, Frostig, Hawkins, Johnson, Leary,
  Maclaurin, Necula, Paszke, Vander{P}las, Wanderman-{M}ilne, and
  Zhang]{bradbury-2018-jax}
J.~Bradbury, R.~Frostig, P.~Hawkins, M.~J. Johnson, C.~Leary, D.~Maclaurin,
  G.~Necula, A.~Paszke, J.~Vander{P}las, S.~Wanderman-{M}ilne, and Q.~Zhang.
\newblock {JAX}: composable transformations of {P}ython+{N}um{P}y programs,
  2018.
\newblock URL \url{https://github.com/google/jax}.

\bibitem[Brockman et~al.(2016)Brockman, Cheung, Pettersson, Schneider,
  Schulman, Tang, and Zaremba]{brockman-2016-openai}
G.~Brockman, V.~Cheung, L.~Pettersson, J.~Schneider, J.~Schulman, J.~Tang, and
  W.~Zaremba.
\newblock Open{AI} {G}ym.
\newblock \emph{CoRR}, abs/1606.01540, 2016.
\newblock URL \url{https://arxiv.org/abs/1606.01540}.

\bibitem[Gendron et~al.(2016)Gendron, Hanafi, and
  Todosijević]{gendron-2016-ilph}
B.~Gendron, S.~Hanafi, and R.~Todosijević.
\newblock An efficient matheuristic for the multicommodity fixed-charge network
  design problem.
\newblock \emph{IFAC-PapersOnLine}, 49\penalty0 (12):\penalty0 117--120, 2016.
\newblock URL \url{https://doi.org/10.1016/j.ifacol.2016.07.560}.

\bibitem[Godwin et~al.(2020)Godwin, Keck, Battaglia, Bapst, Kipf, Li,
  Stachenfeld, Veli\v{c}kovi\'{c}, and Sanchez-Gonzalez]{godwin-2020-jraph}
J.~Godwin, T.~Keck, P.~Battaglia, V.~Bapst, T.~Kipf, Y.~Li, K.~Stachenfeld,
  P.~Veli\v{c}kovi\'{c}, and A.~Sanchez-Gonzalez.
\newblock {J}raph: a library for graph neural networks in {JAX}, 2020.
\newblock URL \url{https://github.com/deepmind/jraph}.

\bibitem[Hagberg et~al.(2008)Hagberg, Schult, and Swart]{hagberg-2008-networkx}
A.~A. Hagberg, D.~A. Schult, and P.~J. Swart.
\newblock Exploring network structure, dynamics, and function using {NetworkX}.
\newblock In \emph{Proceedings of the 7th Python in Science Conference}, pages
  11 -- 15, 2008.
\newblock URL \url{https://www.osti.gov/biblio/960616}.

\bibitem[Klein and Randi{\'{c}}(1993)]{klein-1993-resistance}
D.~J. Klein and M.~Randi{\'{c}}.
\newblock Resistance distance.
\newblock \emph{Journal of Mathematical Chemistry}, 12\penalty0 (1):\penalty0
  81--95, Dec. 1993.
\newblock URL \url{https://doi.org/10.1007/bf01164627}.

\bibitem[Mazyavkina et~al.(2021)Mazyavkina, Sviridov, Ivanov, and
  Burnaev]{mazyavkina-2021-rlco}
N.~Mazyavkina, S.~Sviridov, S.~Ivanov, and E.~Burnaev.
\newblock Reinforcement learning for combinatorial optimization: {A} survey.
\newblock \emph{Computers \& Operations Research}, 134:\penalty0 105400, 2021.
\newblock URL \url{https://doi.org/10.1016/j.cor.2021.105400}.

\bibitem[Nazari et~al.(2018)Nazari, Oroojlooy, Snyder, and
  Tak{\'{a}}c]{nazari-2018-vrp}
M.~Nazari, A.~Oroojlooy, L.~V. Snyder, and M.~Tak{\'{a}}c.
\newblock Reinforcement learning for solving the vehicle routing problem.
\newblock In \emph{Neural Information Processing Systems (NeurIPS)},
  volume~31, pages 9861--9871, 2018.
\newblock URL
  \url{https://proceedings.neurips.cc/paper/2018/hash/9fb4651c05b2ed70fba5afe0b039a550-Abstract.html}.

\bibitem[Pan et~al.(2022)Pan, G{\"{u}}rtler, Neitz, and
  Sch{\"{o}}lkopf]{pan-2022-direct}
H.~Pan, N.~G{\"{u}}rtler, A.~Neitz, and B.~Sch{\"{o}}lkopf.
\newblock Direct advantage estimation.
\newblock In \emph{Neural Information Processing Systems (NeurIPS)},
  volume~35, pages 11869--11880, 2022.
\newblock URL
  \url{https://papers.nips.cc/paper_files/paper/2022/hash/4d893f766ab60e5337659b9e71883af4-Abstract-Conference.html}.

\bibitem[Plappert et~al.(2018)Plappert, Andrychowicz, Ray, McGrew, Baker,
  Powell, Schneider, Tobin, Chociej, Welinder, Kumar, and
  Zaremba]{plappert-2018-multi}
M.~Plappert, M.~Andrychowicz, A.~Ray, B.~McGrew, B.~Baker, G.~Powell,
  J.~Schneider, J.~Tobin, M.~Chociej, P.~Welinder, V.~Kumar, and W.~Zaremba.
\newblock Multi-goal reinforcement learning: Challenging robotics environments
  and request for research.
\newblock \emph{CoRR}, abs/1802.09464, 2018.
\newblock URL \url{https://arxiv.org/abs/1802.09464}.

\bibitem[Qin et~al.(2021)Qin, Zhu, and Ye]{qin-2021-ridehailing}
Z.~T. Qin, H.~Zhu, and J.~Ye.
\newblock Reinforcement learning for ridesharing: {A} survey.
\newblock In \emph{24th {IEEE} International Intelligent Transportation Systems
  Conference (ITSC)}, pages 2447--2454. {IEEE}, 2021.
\newblock URL \url{https://doi.org/10.1109/ITSC48978.2021.9564924}.

\bibitem[Schulman et~al.(2017)Schulman, Wolski, Dhariwal, Radford, and
  Klimov]{schulman-2017-proximal}
J.~Schulman, F.~Wolski, P.~Dhariwal, A.~Radford, and O.~Klimov.
\newblock Proximal policy optimization algorithms.
\newblock \emph{CoRR}, abs/1707.06347, 2017.
\newblock URL \url{https://arxiv.org/abs/1707.06347}.

\end{thebibliography}

\newpage
\appendix
\counterwithin{table}{section}
\counterwithin{figure}{section}

\begin{center}
    {\LARGE\bfseries Supplementary Material}%\\[3mm]
\end{center}
\vspace{1em}

\section{Middle-mile MDP details}\label{sec:mdp}
The main contribution of this work is our open-source implementation of the middle-mile logistics MDP, available online at \url{https://github.com/google-research/laurel}.
The environment has an OpenAI Gym~\cite{brockman-2016-openai}-like interface.
As such, its behavior is encoded in the initialization (\texttt{reset}) and dynamics (\texttt{step}) functions.

\subsection{Initialization}\label{sec:initialization}
The middle-mile state (a time-expanded logistics network with parcels) is created in 5 steps:
\begin{enumerate}
    \item Sampling a static (non-expanded) logistics network.
    \item Creating the expanded network by sampling individual trucks.
    \item Pruning the time-expanded network (removing skippable nodes).
    \item Populating the network with parcels by sampling parcel routes.
    \item Pruning the network by removing nodes and edges that are irrelevant to the parcels.
\end{enumerate}

\begin{wrapfigure}{r}{0.3\textwidth}
  \vspace{-1em}
  \begin{center}
    \includegraphics[width=0.3\textwidth]{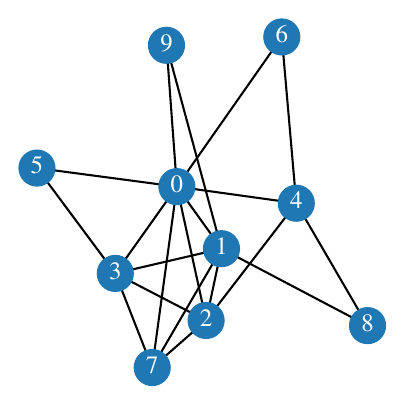}
  \end{center}
  \caption{Static logistics network with 10 hubs}
  \label{fig:static-net}
  \vspace{-1em}
\end{wrapfigure}
There are many algorithms for sampling a random network.
A realistic logistics network includes some major (highly connected) hubs, but a larger number of minor (less connected) hubs.
We generate the static network using the extended Barabási–Albert algorithm~\cite{albert-2000-topology}, as implemented in NetworkX~\cite{hagberg-2008-networkx}, with parameters \(m = 2\), \(p = 0.2\), and \(q = 0\).
A 10-hub network sampled in this way is shown in Fig.~\ref{fig:static-net}.
It can be seen that some nodes have a very high degree, while others have a low degree, as is desired.

\begin{figure}[t]
    \centering
    \includegraphics{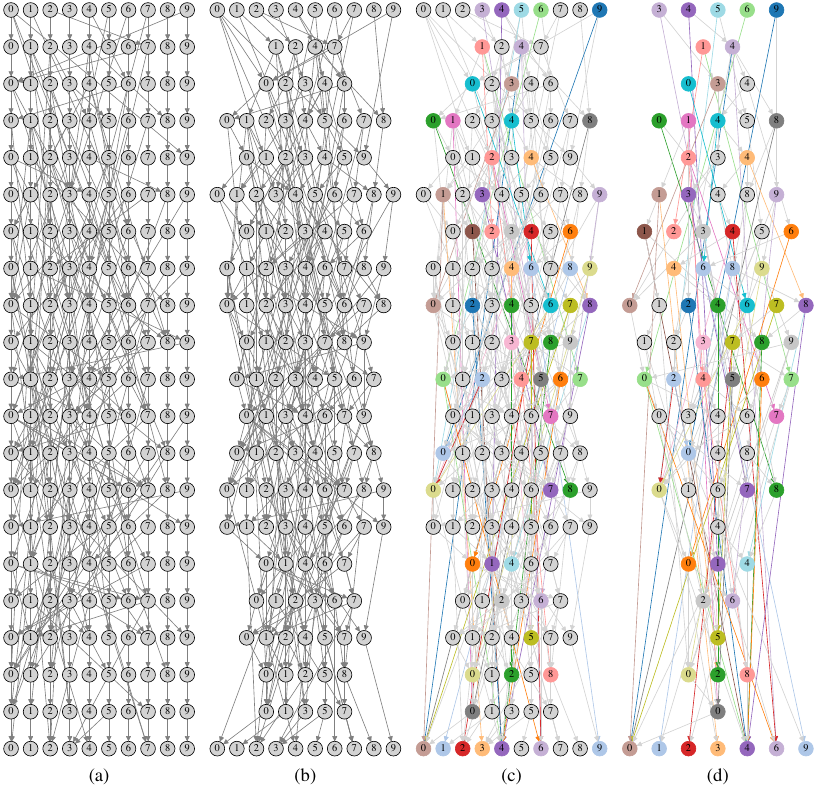}
    \caption{Time-expanded networks after stages 2 to 5 of the MDP initialization procedure. As in Fig.~\ref{fig:time-expanded}, time increases downward. (a) The static 10-hub logistics network from Fig.~\ref{fig:static-net} is expanded by sampling trucks for 20 time steps. (b) Skippable nodes are pruned from the network. (c) The network is populated with 50 parcels, shown in color. The colored edges are not trucks but go from a parcel's start to its goal node. (d) Nodes and edges irrelevant to the parcels are pruned, and newly skippable nodes are again removed.}
    \label{fig:pruning}
    \vspace{-1em}
\end{figure}

In the next step, the network is expanded in time.
We fix the number of trucks leaving at each time step to be equal to the number of hubs (though our interface allows this number to be specified independently).
The individual trucks leaving at a given time are now sampled by selecting an edge from the non-expanded graph and sampling the duration for this truck uniformly from 1 time step to a given maximum duration.
The edges are not sampled uniformly from all connections in the static graph, but connections between major hubs are sampled with a higher probability to make the truck schedule more realistic.
To achieve this, the truck edges are sampled (without replacement) from a Boltzmann distribution:
\begin{equation*}
    p(e) \propto \exp\left\{\beta_1 (\operatorname{deg}(a) + \operatorname{deg}(b))\right\},
\end{equation*}
where \(e\) is an edge connecting hubs \(a\) and \(b\), \(\beta_1\) is an inverse temperature coefficient (we use \(\beta_1 = 0.01\)), and \(\operatorname{deg}\) is the number of neighbors (the degree) of a node. In Fig.~\ref{fig:pruning}(a), a 20-time step expansion of this form is shown for the network from Fig.~\ref{fig:static-net}.
Here, we set the maximum truck duration to 10 time steps.
As in Fig.~\ref{fig:time-expanded}, time increases downwards.
It can be seen that highly connected hubs (such as hub 2) receive much more traffic than less connected hubs (such as hub 9).
Each truck in the network also gets assigned a capacity, which we sample uniformly from the interval \([0, 1]\).

This time-expanded network contains some nodes that have only one incoming and one outgoing truck.
These nodes are not interesting for decision-making, so we remove them for simplicity and replace them with longer edges connecting their parent and child nodes (more details in Sec.~\ref{sec:pruning}).
The pruned result is shown in Fig.~\ref{fig:pruning}(b).

The next step is to add parcels to the network.
We need to make sure that the middle-mile routing problem is solvable, so we can't sample parcel start and goal nodes, or parcel weights, arbitrarily.
Instead, we sample whole parcel routes and keep the start and end nodes as part of the MDP state.
The parcel weights are first sampled from a truncated Pareto (power-law) distribution using rejection sampling to keep the weights below a given threshold.
In our experiments, we use a Pareto-scale parameter (minimum parcel weight) of \(m = 0.01\), a maximum parcel weight of \(1\) (the maximum truck capacity), and a Pareto-shape parameter of \(\alpha = 0.1\).
The power-law distribution of weights is chosen to be realistic; there are many lightweight parcels and few very heavy parcels.

The parcel routes of heavier parcels are sampled first to make it easier to place them in the network.
The initial hub \(h\) for a given parcel is sampled from a Boltzmann distribution encouraging low-degree (minor) hubs: \(p(h) \propto \exp\left\{-\beta_2 \operatorname{deg}(h)\right\}\), where we use \(\beta_2 = 0.1\).
The initial time for a parcel is sampled uniformly from all times in \(\{1, \dots, T - L\}\) that contain the sampled hub \(h\).
Here, \(T\) is the number of time steps of the expanded network, and \(L\) is the mean route length (a free parameter).
The rest of the parcel route is sampled by moving the parcel down in time along trucks that have sufficient capacity.
When sampling transitions, we again use a Boltzmann distribution to encourage routes that take the parcel further from its initial hub.
Thus, a truck \(t\) arriving at hub \(g\) is sampled according to \(p(t) \propto \exp\left\{\beta_3 \operatorname{dist}(h, g)\right\}\), where \(\operatorname{dist}(h, g)\) is the resistance distance\footnote{See Sec.~\ref{sec:feature-graphs} for details.} of \(g\) from the initial hub \(h\) (not necessarily the sender of truck \(t\)), and \(\beta_3\) is another inverse temperature coefficient (we use \(\beta_3 = 0.1\)).

After each step, we have to decide whether to end the parcel route here.
On average, the parcel routes should have a length of \(L\) time steps (the mean route length parameter mentioned above).
We can achieve this by sampling a Bernoulli random variate \(\gamma_t \sim \mathcal{B}(1/L)\) after each time step.
Thus, when \(\gamma_t = 1\), the route terminates after \(t\) time steps.
This way, the expected route length \(N\) is:
\begin{equation*}
    \mathbb E N = \mathbb E \sum_{t = 1}^\infty [N \geq t] = \mathbb E \sum_{t = 1}^\infty \prod_{i = 1}^{t-1} (1 - \gamma_i) = \sum_{t = 1}^\infty \prod_{i = 1}^{t-1} \left(1 - \frac{1}{L}\right) = \sum_{t = 0}^\infty \left(1 - \frac{1}{L}\right)^t = L,
\end{equation*}
where \([\cdot]\) is the Iverson bracket, and we have made use of the independence of the termination variables \(\gamma_t\).
In our case, this expectation only holds approximately for two reasons.
Firstly, there is only a finite number \(T\) of time steps, so the geometric series is truncated after \(T\) terms.
Secondly, we can't terminate at every time step.
If a truck duration takes longer than one time step, we can still only terminate the parcel route once the truck has reached its destination.
Thus, after taking the \(k\)\textsuperscript{th} truck, we actually sample a binomial random variate \(\gamma_k \sim \operatorname{Bin}(\ell_k, 1/L)\), where \(\ell_k\) is the truck duration, and terminate the route if \(\gamma_k > 0\).

Once a complete parcel route has been sampled, we check whether the end hub is different from the start hub.
If this is not the case, for example because all sampled trucks were virtual, we reduce the parcel weight by 10\% to make it easier to place and try again.
If the parcel route still routes the parcel to its start hub after 50 retries, we keep this suboptimal route.
A logistics network populated with parcels sampled in this fashion is shown in Fig.~\ref{fig:pruning}(c).
Finally, the network is pruned again, this time removing all nodes and edges that are irrelevant to the routing problem defined by the parcels and their goals (see Sec.~\ref{sec:pruning}).
The final result is shown in Fig.~\ref{fig:pruning}(d).
Almost all of the parameters mentioned above can be customized by passing the desired value to the MDP initialization function.

To fully specify the MDP state as a graph, each edge has an associated feature vector.
There are three types of edges in the state.
First, we have trucks.
Each truck is represented by two edges, one going from the start hub to the destination hub (forward in time), and one going the other way (backward in time).
Both edges have the truck capacity as a feature.
The second type of edge is the virtual truck, which is how we represent the option to stay at the same location for one time step.
These virtual trucks don't have capacity constraints, and are also represented by one edge forward in time, and one backward in time.
Finally, there are parcels, which are also represented by two edges connecting the start and goal hub, going in opposite directions in time.
Parcels additionally have the parcel weight as a feature.
We represent all edges using vectors of the form
\begin{equation*}
    ([\text{truck}\ \oplus], [\text{truck}\ \ominus], [\text{parcel}\ \oplus], [\text{parcel}\ \ominus], [\text{virtual}\ \oplus], [\text{virtual}\ \ominus], \text{truck capacity}, \text{parcel weight}),
\end{equation*}
where the \([\cdot]\) fields are a one-hot encoding of the type of edge (\(\oplus\) meaning forward in time and \(\ominus\) meaning backward in time), and the last two fields are zero when not applicable.
The node features in our state representation are of the form (hub, time).
We go into more detail on the node and edge features in Sec.~\ref{sec:feature-graphs}.

\subsection{Dynamics}
The MDP dynamics, which are illustrated in Fig.~\ref{fig:time-expanded}, are relatively straightforward.
Given an MDP state, a parcel, and a truck or virtual truck starting at the parcel's location (in time and space), the \texttt{step} function returns a new state in which the parcel has been moved to the truck's destination (in time and space) and the truck's capacity has decreased by the parcel's weight.
If this transition results in a successful delivery of a parcel to its goal node, the function also returns this information, and the parcel is removed from the state.
The parcel is also removed if the transition moves it to any node that lies later in time than the goal node, as this parcel cannot possibly be delivered anymore.
Optionally, the state is also pruned after each transition (see Sec.~\ref{sec:step-pruning}).
In our experiments, we choose not to do this for performance reasons.

\section{State pruning}\label{sec:pruning}
As discussed in the main text, the main challenge of the middle-mile problem is the size of the time-expanded graphs.
Not only are larger problems inherently more difficult to solve, but large graphs also present a problem to our proposed GNN-based method.
Conceptually, the most straightforward way to decrease the graph size is to \textit{prune} the graph, by removing parts that are irrelevant to finding a good solution.
We have implemented three different kinds of pruning methods: skip pruning, parcel pruning, and step pruning, which we go over in the following subsections.
In Sec.~\ref{sec:hamlet}, we discuss the benefits and drawbacks of pruning.

\subsection{Skip pruning}
We call the operation in step 3 of the MDP initialization (Sec.~\ref{sec:initialization}) \textit{skip pruning}, illustrated as the transition from (a) to (b) in Fig.~\ref{fig:pruning}.
Some nodes in the MDP state have only one parent node and one child node, only counting (virtual) truck edges that go forward in time, and are neither the current location nor the goal node of a parcel.
Thus, these nodes are irrelevant to the decision-making problem: what goes in from above must come out below.
Skip pruning removes these ``skippable'' nodes.
First, all nodes of this kind are identified in the state.
Then, these nodes are grouped together in case some of them form longer chains of skippable nodes.
Finally, the groups are removed and replaced by a single new connection.
This connection is a truck whose capacity is the minimum of the group's trucks' capacities (ignoring virtual trucks).
If all trucks in a group are virtual, then the new connection is also a virtual truck.

It is important to keep track of which nodes are removed.
Let's say skip pruning removes a node belonging to hub E of the example in Fig.~\ref{fig:middle-mile} and replaces it with a direct connection from B to D.
If the solver selects this connection as part of a parcel's solution, that is not valid, because the connection does not actually exist.
Thus, our implementation returns a description of each new connection in terms of the individual nodes that are being skipped.

\subsection{Parcel pruning}\label{sec:parcel-pruning}
We only care about those parts of a logistics network that are relevant to the problem of routing the parcels it contains.
For instance, if all parcel locations and goals are located in Spain, we don't care about the truck schedule in Korea.
Similarly, if all parcels should be delivered within the next week, we don't care about next month (or last year).
Given a parcel, the \textit{parcel pruning} operation removes parts of the state that cannot be part of a valid solution for this parcel by considering reachability.
All nodes that are part of a valid parcel route must be reachable, both from the parcel's start node and from the parcel's goal node, via trucks with sufficient capacity.
Our implementation finds these parts of the state by starting at the parcel's start node and iteratively expanding all valid trucks going forward in time until half the duration toward the goal is covered.
Then, the same expansion is done starting at the goal node, and moving backward in time until the two expansion trees meet in the middle.
Finally, only those nodes and edges that are reachable from both the top and the bottom are kept.
This two-way procedure is computationally more efficient than only starting at the top and expanding downward until reaching the goal.

When done for a single parcel, this pruning already solves the routing problem (which, as mentioned in the main text, reduces to a shortest-path problem in this case).
However, with several parcels, this is not the case, as the reachability check assumes all trucks are empty.
Parcel pruning can be useful in several different ways.
For one thing, it can be used to reduce the size of the state.
The procedure is simply repeated for all parcels in the network and the reduced state is built from the union of all reachable nodes and edges for the individual parcels.
This is what we do in step 5 of the initialization (Sec.~\ref{sec:initialization}), followed by a second application of skip pruning.
An illustration of this is the transformation from (c) to (d) in Fig.~\ref{fig:pruning}.

Parcel pruning can also be useful when applied to only a single parcel, however.
For example, when deciding which truck to use next for a specific parcel, pruning the state such that only trucks relevant to the given parcel remain can be very useful.
We go into more detail on this strategy in Sec.~\ref{sec:hamlet}, where we explain why we choose not to do this in practice.

\subsection{Step pruning}\label{sec:step-pruning}
\begin{figure}
    \centering
    \includegraphics{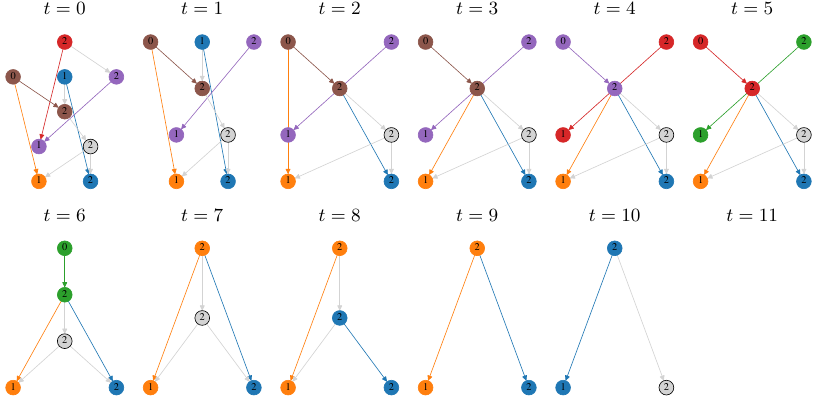}
    \caption{Step pruning removes only those parts of the state graph that are affected by an MDP transition (moving a parcel). In this 3-hub, 5-time step, 5-parcel problem, all parcels are routed correctly, and the state shrinks over the course of the episode as a result of step pruning. The colored arrows here denote parcel edges (from a parcel's location to its goal), but they might block the view of truck edges lying underneath. A parcel's color is not fixed and might change in this plot.}
    \label{fig:step-pruning}
\end{figure}

The third method of pruning the state, \textit{step pruning}, is about maintaining a minimal state during rollouts.
Given a fully pruned state, for example that of a freshly initialized MDP, it is not necessary to repeat the whole costly pruning operation after the first transition.
In this transition, only a single parcel moves, so it should be enough to focus on that region of the state graph.
This is possible by starting at the node where the parcel was before the transition, and again considering reachability.
If this node is not reachable anymore, i.e. it was only relevant to the parcel that was just moved, we can iteratively expand all neighbors from this node that are not reachable in a different way.
All these nodes can be removed from the state, and we again get a minimal state representation.
However, even though this operation is computationally cheaper than running the parcel pruning for all parcels, it is not for free.
In our experiments, we decided against using it; more details in Sec.~\ref{sec:hamlet}.
Step pruning is illustrated on a small MDP in Fig.~\ref{fig:step-pruning}, where it can be seen to maintain a minimal state.

\subsection{To prune, or not to prune?}\label{sec:hamlet}
\begin{figure}[t]
    \centering
    \includegraphics{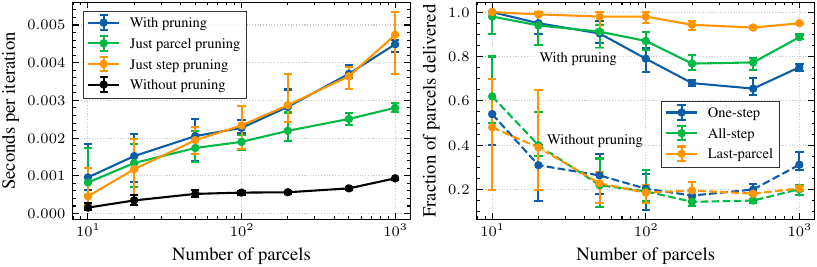}
    \caption{Left: Comparison of times taken by a complete MDP transition. The call to \texttt{get\_actions} can include parcel pruning and the call to \texttt{step} can include step pruning. Right: Comparing the performance of the uniform random policy in an environment with pruning (solid lines) and without pruning (dashed lines) using different high-level routing strategies. Both plots show the minimum, the maximum, and the mean values of 5 experiments with different random seeds.}
    \label{fig:random-pruning}
\end{figure}

The main drawback of pruning is that it takes time.
In the left plot of Fig.~\ref{fig:random-pruning}, we analyze how the time that it takes to do a full MDP transition (on an M1 Pro MacBook) depends on the amount of pruning.
The MDP here is a 10-hub, 50-time step setup as we use in our main experiments (see Sec.~\ref{sec:experiments}).
The policy is uniformly random (a single call to \texttt{np.random.choice}) in all cases, and this call is also included in the measured time.
It can be seen that the pruning operations take the most amount of time.
Here, parcel pruning refers to pruning all nodes and edges that are irrelevant for the given parcel at decision time, used in \texttt{get\_actions}.
It can be seen that both pruning operations considerably slow down the simulation.
Interestingly, however, it seems that the more expensive step pruning operation improves the state representation so much that parcel pruning comes at no additional computational cost.

Of course, pruning can also bring benefits.
As mentioned above, the parcel pruning operation in \texttt{get\_actions} removes all actions (available trucks) that don't lead to the parcel's goal.
In the right plot of Fig.~\ref{fig:random-pruning}, we compare three different high-level routing strategies employed with and without pruning: one-step routing, all-step routing, and last-parcel routing.
These strategies are explained in Sec.~\ref{sec:hierarchy}.
It can be seen that pruning improves performance greatly.
In our experiments, we do not use any pruning (except during initialization, see Sec.~\ref{sec:initialization}) to save on computational resources.
However, it is reasonable to assume that a policy trained in an environment without pruning would still work well with pruning at inference time.

\section{Feature graphs}\label{sec:feature-graphs}
The pruning techniques discussed in Sec.~\ref{sec:pruning} are intended to make the state smaller and more manageable.
However, unless the problem at hand is particularly small, the fully pruned states are still too large to reasonably process with message-passing graph neural networks.
For this reason, we don't use the full state as the input to the actor and critic networks.
Instead, given the parcel that is to be routed next, we extract a smaller graph from the state that is specific to this parcel.
To construct these ``feature graphs'', we start with only the start node, the goal node, and the parcel edge connecting them.
We can then iteratively expand all nodes in the current graph by adding all their incoming and outgoing edges and the corresponding sending and receiving nodes.
The number \(K\) of these expansion steps controls the size of the feature graph: the final graph contains all nodes within a radius of \(K\) steps from the parcel start and goal nodes.
After this extraction procedure, we also add new virtual truck connections between any nodes in the feature graph belonging to the same hub whose virtual connections have been pruned.
This is necessary because the parcel's goal node can be much later in time than its start node: with low \(K\), the two parts of the feature graph can be disconnected except for the parcel connections.
As it is always possible to stay at a hub for any number of time steps, these additional virtual connections are intended to connect the top and bottom parts of the feature graph.

\begin{figure}[t]
    \centering
    \includegraphics{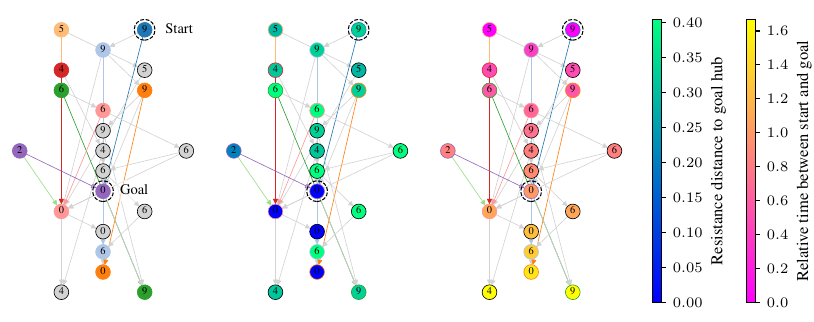}
    \caption{A 2-step feature graph corresponding to the top right parcel in Fig.~\ref{fig:pruning}(d). In addition to the usual representation (left graph), we also show the two new node features: resistance distance to goal hub (middle graph), and relative time between start and goal (right graph).}
    \label{fig:feature-graphs}
\end{figure}

In the left of Fig.~\ref{fig:feature-graphs}, a feature graph (\(K = 2\)) corresponding to the top right parcel of Fig.~\ref{fig:pruning}(d) is shown.
This graph is clearly easier to process than the large state graph.
However, the feature graph no longer has all the information contained in the state, so any middle-mile solver based on feature graphs will necessarily only yield approximate solutions.
As exact solution methods are infeasible anyway, this is not particularly problematic.
To improve the quality of these solutions, we can include some of the information that has been lost in the form of new node and edge features.
In particular, the node features are of the form (resistance distance to goal hub, relative time between start and goal).
For the example feature graph, both these node features are illustrated in Fig.~\ref{fig:feature-graphs}.
The \textit{resistance distance} \cite{klein-1993-resistance} between two nodes in an undirected graph is obtained by assigning an electrical resistance to each edge of the graph and solving for the equivalent resistance between the two nodes.
In our case, the undirected graph is the static logistics network (shown in Fig.~\ref{fig:static-net} for this example), and the resistances are assigned according to how frequently trucks are sampled from an edge.
More concretely, we set the electrical conductance (inverse resistance) of an edge connecting the hubs \(a\) and \(b\) equal to \(\beta_1 (\operatorname{deg}(a) + \operatorname{deg}(b))\) (see Sec.~\ref{sec:initialization} for details).
A node's relative time \(\tilde t\) between the start node \(s\) and the goal node \(g\) is defined as: 
\begin{equation*}
    \tilde t =
    \begin{cases}
        (t - t_s) / (t_g - t_s) & \text{if}\ t_g > t_s\\
        0 & \text{otherwise,}
    \end{cases}
\end{equation*}
where \(t\) is the node's time and \(t_s\) and \(t_g\) are the start and goal nodes' times, respectively.

In addition to the new node features, we also add three new edge features to the feature graph in addition to the ones described at the end of Sec.~\ref{sec:initialization}:
\begin{enumerate}
    \item A binary feature encoding whether a parcel edge corresponds to the parcel that is currently being routed,
    \item A binary feature encoding whether a truck edge is one of the available trucks (actions) for this parcel, and
    \item A \textit{phantom parcel weight}, assigned to real trucks, that contains information about those parcels that are cut from the feature graph, but that might also use this truck.
\end{enumerate}
This is important information, as consolidating the needs of different parcels is at the core of the middle-mile problem.
The phantom parcel weight of a truck can be thought of as the expected value of the truck's capacity taken up by all parcels not included in the feature graph, given that each of them is routed using a uniformly random policy.
These phantom parcel weights are computed using the parcel pruning operation (Sec.~\ref{sec:parcel-pruning}).
The parcel pruning operation done during the MDP state initialization (Sec.~\ref{sec:initialization}) already records to which parcels a given edge is possibly relevant.
We now go over all edges in the feature graph and collect these parcels.
Each of these parcels possibly interferes with the capacities of trucks in the feature graph, and we would like to include them as phantom parcels.
To do this for one parcel, we distribute the parcel's weight across the edges of the parcel-pruned graph.
At the parcel's start node, its phantom weight is equal to its real weight, \(w\).
If there are \(n\) available trucks from this node, each truck receives a phantom weight of \(w'\ \dot=\ w/n\).
If there are now \(n'\) available trucks from one of the \(n\) trucks' receiving nodes, each of these gets assigned a phantom weight of \(w'/n'\).
Finally, to get the phantom parcel weight associated with an edge in the feature graph, all of the phantom weights computed for this edge due to the different parcels are added together.

As discussed in Sec.~\ref{sec:hamlet}, the parcel pruning operation takes some time.
In our experiments, we therefore don't include phantom weights in the feature graphs for performance reasons.

\section{Further details}\label{sec:experiments}
\subsection{High-level routing strategies}\label{sec:hierarchy}
In the standard goal-conditioned reinforcement learning setting (e.g.~\cite{plappert-2018-multi}), there is only a single object with a goal.
Our setting is slightly different because we have a large number of objects (parcels), each of which has its own goal.
If we were only satisfied once all parcels have reached their respective goals, then these settings would be equivalent, with our goal being the state where all parcels are delivered.
However, we aim to deliver as many parcels as possible, and so the multi-object goal-conditioned setting is more appropriate.

Another way to look at this setting is as hierarchical goal-conditioned RL.
Each low-level decision is concerned with the problem of routing a single parcel for one step.
The high-level decision is about choosing which parcel to route next.
We don't attempt to learn a high-level policy but only learn the low-level policy.
However, we did experiment with three different high-level routing strategies: one-step routing, all-step routing, and last-parcel routing.
In one-step routing, the earliest parcel in the network is always routed first.
The rationale behind this strategy is that, in a real-time setting, this is the most urgent decision to make.
After this parcel has been routed for one step, its location in time has changed.
The next parcel to be routed will again be the earliest one in the network, which is not necessarily the same parcel.
In all-step routing, it is also the earliest parcel that is selected, but here it is routed for as many steps as it takes to reach a node with a time greater or equal than its goal node (this includes the case of a successful delivery).
In last-step routing, it is always the latest parcel in the network that is delivered first.
Here, there is no difference between ``one-step'' and ``all-step'' strategies, because the latest parcel in the network necessarily stays the latest one after each transition.
These three high-level routing strategies are compared in the right plot of Fig.~\ref{fig:random-pruning}, where the low-level policy is uniformly random.
If only those actions that lead to the parcel's goal are available (``with parcel pruning''), then routing later parcels first, which is what both the last-parcel and all-step routing strategies do, gives an advantage.
This is because earlier parcels will automatically react to these decisions, as those trucks that only lead to the goal via later trucks that are necessary for later parcels will not be included in the available actions after these later trucks have been filled.
As we don't employ parcel pruning during episodes, we always use the one-step routing strategy.

\subsection{Hyperparameters}
Unless stated otherwise, we used the following hyperparameters in our experiments:

\begin{table}[h]
    \centering
    \begin{tabular}{lr}
    \toprule
    Parameter & Value\\ 
    \midrule 
    Number of hubs \(H\) & \(10\)\\
    Number of time steps \(T\) & \(50\)\\
    Number of trucks per time step & \(H\)\\
    Maximum truck duration [time steps] & \(5\)\\
    Number of parcels & \(200\)\\
    Mean route length \(L\) & \(10\)\\
    Boltzmann exploration inverse temperature \(\alpha\) & \(0.1\)\\
    PPO clipping threshold \(\epsilon\) & \(0.2\)\\
    PPO early stopping KL divergence threshold\tablefootnote{The KL threshold was intentionally set to this larger value because we only processed the graphs in small batches of 256, leading to a higher variance in the KL divergence estimate.} & \(0.1\)\\
    SL: Number of rollouts & \(100\)\\
    SL: Number of feature graph steps \(K\) & \(3\)\\
    SL: Number of epochs & \(5\)\\
    RL: Total number of rollouts & \(1000\)\\
    RL: Number of feature graph steps \(K\) & \(2\)\\
    RL: Number of gradient steps per epoch \(N_u\) & \(50\)\\
    RL Linear: Number of rollouts per epoch \(N_r\) & \(1\)\\
    RL Linear: Actor learning rate \(\eta_a\) & \(0.01\)\\
    RL Linear: Critic learning rate \(\eta_c\) & \(0.01\)\\
    RL GNN: Number of rollouts per epoch \(N_r\) & \(5\)\\
    RL GNN: Actor learning rate \(\eta_a\) & \(0.0001\)\\
    RL GNN: Critic learning rate \(\eta_c\) & \(0.001\)\\
    \bottomrule
    \end{tabular}
\end{table}

Certain hyperparameter values not listed here can be found in Sec.~\ref{sec:initialization}.
The following hyperparameters were optimized separately for the linear and GNN versions of the RL method in a grid search: \((\eta_a, \eta_c) \in \{0.0001, 0.001, 0.01\}^2, N_r \in \{1, 2, 5\}, N_u \in \{10, 20, 50\}\).

In all our experiments, we also used a slightly modified version of the middle-mile MDP.
In this version, all trucks have a capacity of \(1\), and all parcels have a weight of \(1\).
Additionally, after the state has been initialized, all trucks that are not part of any parcel route are removed from the network.
These modifications bring out the central hardness of the middle-mile problem even for smaller instances, as trucks can only take a single parcel now.
This version of the MDP is also part of our public implementation and can be accessed by passing the appropriate parameters at initialization.

\end{document}